\documentclass[runningheads]{llncs}
\usepackage{graphicx}
\usepackage{booktabs}
\usepackage[colorinlistoftodos]{todonotes}
\usepackage[inline]{enumitem}
\usepackage{cite}
\usepackage{amsmath,amssymb,amsfonts}
\usepackage{algorithmic}
\usepackage{hyperref}
\usepackage{textcomp}
\usepackage{xcolor}
\usepackage{comment}
\usepackage{pgf-pie} 

\usepackage{chngcntr}
\counterwithout{figure}{chapter}
\counterwithout{table}{chapter}

\begin{document}
\title{Analyzing Race and Country of Citizenship Bias in Wikidata}
%
%
\author{Zaina Shaik, Filip Ilievski, Fred Morstatter} 

\authorrunning{Shaik, Ilievski, and Morstatter}
%
\institute{Information Sciences Institute, University of Southern California}
\maketitle              
\setcounter{footnote}{0}
\setcounter{table}{0}

\begin{abstract}\let\thefootnote\relax\footnotetext{Copyright © 2021 for this paper by its authors. Use permitted under Creative Commons License Attribution 4.0 International (CC BY 4.0).}
As an open and collaborative knowledge graph created by users and bots, it is possible that the knowledge in Wikidata is biased in regards to multiple factors such as gender, race, and country of citizenship.
Previous work has mostly studied the representativeness of Wikidata knowledge in terms of genders of people.
In this paper, we examine the race and citizenship bias in general and in regards to STEM representation for scientists, software developers, and engineers. By comparing Wikidata queries to real-world datasets, we identify the differences in representation to characterize the biases present in Wikidata. Through this analysis, we discovered that there is an overrepresentation of white individuals and those with citizenship in Europe and North America; the rest of the groups are generally underrepresented. Based on these findings, we have found and linked to Wikidata additional data about STEM scientists from the minorities. This data is ready to be inserted into Wikidata with a bot. 
Increasing representation of minority race and country of citizenship groups can create a more accurate portrayal of individuals in STEM.
\end{abstract}
\section{Introduction}
Data is the most powerful when it is accurately represented. A knowledge graph is a collection of knowledge that uses a graph structured data model to represent objects through their attributes and their relationships to other objects. Wikidata\cite{vrandevcic2014wikidata} is an open knowledge graph that contains encyclopedic knowledge, similar to Wikipedia, only in a structured form. Wikidata has been collaboratively created by users and bots. As an open and collaborative knowledge graph created by users and bots, it is possible that the data being inputted is biased in regards to multiple factors such as gender, race, and country of citizenship. Biased data leads to inaccurate representation which is harmful because it can influence the perspective of its viewers.

Personalized knowledge graphs are known to have people bias and algorithm bias \cite{gerritse2020personalized}.
Prior work has typically focused on gender bias, showing that there exists a gender disparity in Wikidata with women being underrepresented \cite{klein2016monitoring}. By including more data representing women, it is possible to decrease some of the gender bias present \cite{klein2016monitoring}.  Data visualizations about gender biases can help bring awareness to the gender bias in Wikidata\footnote{\url{http://datakolektiv.org/app/WDCM_BiasesDashboard}, accessed on July 9, 2021.}. Representation in terms of other demographic characteristics, such as race and country of citizenship, is also necessary for an accurate portrayal of knowledge. However, to our knowledge, these have not been studied in previous work.

In this paper,
we tackle the representation of race and citizenship in Wikidata. 
For this purpose, we acquire representative datasets with general statistics on race and country of citizenship. We compute the corresponding statistics for people in Wikidata, by using existing tools, such as the Knowledge Graph Toolkit \cite{ilievski2020kgtk} and SPARQL\footnote{\url{https://www.w3.org/TR/sparql11-query/}}. We measure bias by comparing the real-world data statistics to the statistics in Wikidata. 
Besides a general representativeness, representation in Wikidata can be different based on the groups of people. Careers in STEM already have a known lack of representation with some non-European countries and ethnicities\footnote{\url{https://www.pewresearch.org/science/2021/04/01/stem-jobs-see-uneven-progress-in-increasing-gender-racial-and-ethnic-diversity/}}. We explore whether there is also a lack of representation of people in STEM careers from non-European countries in Wikidata. We choose to find queries for scientists, software developers, and engineers that have an ethnicity and country of citizenship property recorded. We grouped them by races and continents and compared to external datasets to see if there were groups of people overrepresented and underrepresented. Our analysis shows that for most subcategories of STEM careers, there is an overrepresentation of individuals of the white race and from European and North American countries.

Similar to~\cite{gerritse2020personalized}, the bias can be alleviated by adding 
more data from groups of races and countries of citizenship where representation was lacking into the appropriate spaces of Wikidata. For this purpose, we found and linked to Wikidata a table with STEM scientists that come from minority groups, which could be easily inserted into Wikidata with a bot. By increasing the representation present in Wikidata, we can help create a better perspective of the backgrounds of individuals with STEM careers and in general.


The paper is structured as follows. In Section~\ref{sec:background} we present prior studies of bias in Wikimedia knowledge bases, and we list our research questions and hypotheses. Our method for measuring bias in Wikidata is presented in Section~\ref{sec:method}. We report and discuss our results in Section~\ref{sec:results}. We conclude the paper in Section~\ref{sec:conclusion}.

\section{Background}
\label{sec:background}

\subsection{Past Studies of Bias in Wikimedia Knowledge Bases}


By comparing the gender disparity in Wikidata to the gender disparity in the real world, previous work has been able to detect gender bias \cite{klein2016monitoring}. They introduced the Wikidata Human Gender Indicators (WHGI), a dataset,  to show the changing representation of women \cite{klein2016monitoring}. There is a bias present towards the English language in Wikidata's content \cite{kaffee2018language}. Wikidata also contains social bias related to professions by reinforcing  gender, religion, ethnicity and nationality stereotypes \cite{fisher2019measuring}. To our knowledge, no prior work has investigated the bias of Wikidata in terms of race and country of citizenship.

\subsection{Research Questions and Hypotheses}
We looked at three main research questions. First, \textit{how biased is Wikidata in terms of race and country of citizenship representation?} We hypothesize that bias exists towards the white race and European countries because only a few Wikidata editors contribute most of the Wikidata edits \cite{sarasua2019evolution} and there has been evidence of gender bias skewing towards men. Second, \textit{how biased is Wikidata in terms of race and country of citizenship representation for scientists, software developers, and engineers?} We hypothesize that bias exists towards the white race and European countries due to the existing bias in STEM careers. Finally, we investigate \textit{how we can alleviate the bias by adding more racial and country of citizenship representation}. 
Real-world catalogues with scientists from minority groups can be found publicly online. We hypothesize that these could be linked to Wikidata in a semi-automatic manner, and inserted into Wikidata with a bot, in order to
instill more representation and create a better view of underrepresented communities. 

\section{Method}
\label{sec:method}

We define bias towards a demographic group (e.g., race or continent group) in Wikidata as a discrepancy between its representation in Wikidata and in a representative external source. A group can be \textit{overrepresentated}, when its percentage in Wikidata is higher than that in the external source, or \textit{underrepresented}, in case Wikidata contains a lower percentage of people from that group in comparison to the real-world source. The representation of underrepresented groups can be improved by injecting catalogues with people that belong to these categories with a minority bot. We next describe how we obtain the datasets to perform this comparison and adjustment of bias.



\subsection{Measuring Bias}

To measure biases in Wikidata, we compare its data to representative real-world datasets.

\textbf{Wikidata} We ran Wikidata queries using SPARQL and KGTK to collect the information already available on Wikidata. Starting by finding datasets to observe racial and ethnic bias, we looked at the 50 most frequent ethnicities in Wikidata in general found by searching for individuals with an an Ethnicity property (P172). About 90\% of all results for each query were captured in the top 50 categories. Then, we did the same with scientists, software developers, and engineers who have an ethnicity property in Wikidata. Since Wikidata only contains ethnicity and not race information, we manually grouped the ethnicities into the following race categories: Asian, White, Black, Middle Eastern, Indigenous to America, Hispanic/Latino, and Pacific Islander. We chose these race categories to include the U.S. Office of Management and Budget's minimum five categories and added Middle Eastern and Hispanic/Latino for more specific identification. We used the term indigenous to be more inclusive\footnote{\url{https://indigenousfoundations.arts.ubc.ca/terminology/}}.


To observe country of citizenship bias on Wikidata, we ran Wikidata queries to find the 50 most frequent countries of citizenship in Wikidata in general and for scientists, software developers, and engineers by searching for individuals with a Country of Citizenship property (P27). 
We manually grouped the countries into the following continent categories: Asia, Europe, Middle East, Africa, North America, South America, and Pacific Islands.

\textbf{Real-world data} In order to have a reference point of real-world data available outside of Wikidata, we found external datasets including the population of races in the world and the population of countries in the world. The latter was also manually grouped into the previously stated continent categories. The full list of sources used can be found at \url{https://bit.ly/3AdZ80f}.

Notably, the real-world data represents people that are alive today, while Wikidata captures a time-agnostic set of people, many of which are not alive today. This limitation should be considered in subsequent work, e.g., by excluding people that have a death date from the Wikidata statistics.






\subsection{Increasing Representativeness of Wikidata}

The representativeness of a group can be improved by inserting an external catalogue of people that belong to this group into Wikidata. As a proof of concept, we adapted a catalogue of 112 Black scientists, from \url{https://bit.ly/3lxqgTS}. Each persons is described with their first name, surname, year of birth, year of death, occupation, inventions and accomplishments, and wikipedia link. As the scientists were not linked to Wikidata, we ran a table linking software\footnote{\url{https://github.com/usc-isi-i2/table-linker}} to link the Black scientists dataset to Wikidata. For example, Harold Amos was linked to the Wikidata entity with an identifier \texttt{Q5659918}. Missing entities, like Earl S. Bell, will be assigned new Wikidata identifiers and described with their corresponding information. The linked table can then be injected into Wikidata with a bot, by using the \texttt{pywikibot} Python package.\footnote{\url{https://www.wikidata.org/wiki/Wikidata:Pywikibot_-_Python_3_Tutorial}}

The described procedure of obtaining, linking, and injecting external catalogues with minority groups into Wikidata can be generalized to other groups. We list other catalogues for underrepresented groups, that represent Asian scientists, Hispanic/Latinx scientists, and Middle Eastern scientists in \url{https://bit.ly/3AdZ80f}.


\section{Results}
\label{sec:results}

\textbf{Measuring race bias} The results that report on the race of scientists, software developers, and engineers are shown in Table~\ref{tab:race_bias}. In terms of these STEM categories and overall Wikidata is skewed towards the white race while underrepresenting all other races. While white people make up 17.80\% of the world population, in Wikidata, this percentage is higher than 37.63\%. The representation of white people is consistently higher for STEM careers: white scientists made up 83.95\%, white software developers - 44.08\%, and white engineers - 70.74\%. This validates our hypothesis that Wikidata is biased in terms of race, overrepresenting white people, and underrepresenting all other races. 

\noindent \textbf{Measuring citizenship bias} The results for country of citizenship comparison are in Table~\ref{tab:citizenship_bias}. Wikidata is skewed towards European and North American countries while underrepresenting all other continents. Compared to the world population (Europe: 8.80\%, North America: 5.45\%), Wikidata overrepresents European and North American people. The percentage of Europeans and North Americans in Wikidata is 3-6 times higher in comparison to the real-world statistics. In particular, 57\% of the people in Wikidata have a European country of origin, compared to only 8.80\% in the real world. Similar to the race observation, we see a higher bias for the STEM professions of scientists (71.06\% and 15.43\% for Europe and North America, respectively), software developers (68.09\% and 19.88\%, respectively), and engineers (66.82\% and 19.51\%, respectively).

\noindent \textbf{Injecting minority data into Wikidata}

\begin{table}[t!]
    \centering
    \caption{Percentage of accurate Qnodes matched from Black Scientists dataset to Wikidata.}
    \label{tab:link_acc}
    \begin{tabular}{lll}
    Accuracy & Person & Occupation \\
    Correct & 93.75 & 59.17 \\
    Candidate Ranking Error & 4.47 & 35.00 \\
    Candidate Generation Error & 1.79 & 5.83
    \end{tabular}
\end{table}
The table linking software took information from the dataset and ranked five possible matching Wikidata Qnodes for each person and occupation. A Candidate Ranking Error occured when the software found the correct Qnode, but not as the first option. A Candidate Generation Error refers to when the software did not find the matching Qnode. The linking results of the table linking software were manually validated by one of the authors of this paper. It matched up the table columns to existing entities and properties in Wikidata with a 93.75\% accuracy rate. We started the registration for a MinorityBot to insert individuals from underrepresented race and country groups into Wikidata.

\begin{table}[t!]
    \centering
    \caption{Comparison of race distributions between Wikidata (WD) and real-world (real) data.}
    \label{tab:race_bias}
    \begin{tabular} {p{4cm} r r r r r r r r}
    \hline
    & \multicolumn{2}{c}{Total} & \multicolumn{2}{c}{Scientists} & \multicolumn{2}{c}{Software} & \multicolumn{2}{c}{Engineers} \\
    & WD & real & WD &  & WD &  & WD &  \\ \hline
    White & 37.63 & 17.80 & 83.95 &  & 44.08 &  & 70.74 & \\
    Black & 18.60 & 14.83 & 9.05 &  & 16.95 &  & 14.72 &  \\
    Asian & 39.35 & 31.14 & 1.10 &  & 20.34 &  & 5.97 &  \\
    Middle Eastern & 2.37 & 24.57 & 5.46 &  & 15.25 &  & 6.75 &  \\
    Indigenous to America & 0.82 & 3.71 & 0.00 &  & 1.69 &  & 0.61 &  \\
    Hispanic/Latino & 1.15 & n/a & 0.44 &  & 0.00 &  & 1.15 &  \\
    Pacific Islander & 0.18 & n/a & 0.00 &  & 1.69 &  & 0.00 &  \\
    Other & n/a & 7.95 & n/a &  & n/a &  & n/a &  \\ \hline
    \end{tabular}
\end{table}


\begin{table}[t!]
    \centering
    \caption{Comparison of country of citizenship distributions between Wikidata (WD) and real-world (real) data.}
    \label{tab:citizenship_bias}
    \begin{tabular} {p{4cm} r r r r r r r r}
    \hline
    & \multicolumn{2}{c}{Total} & \multicolumn{2}{c}{Scientists} & \multicolumn{2}{c}{Software} & \multicolumn{2}{c}{Engineers} \\
    & WD & real & WD &  & WD &  & WD &  \\ \hline
    Asia & 18.06 & 55.26 & 7.47 &  & 5.46 &  & 5.36 & \\
    Europe & \textbf{57.10} & \textbf{8.80} & \textbf{71.06} &  & \textbf{68.09} &  & \textbf{66.82} &  \\
    Middle East & 1.97 & 5.55 & 1.30 &  & 1.34 &  & 1.32 &  \\
    Africa & 0.00 & 11.90 & 0.00 &  & 0.00 &  & 0.00 &  \\
    North America & \textbf{16.47} & \textbf{5.45} & \textbf{15.43} &  & \textbf{19.88} &  & \textbf{19.51} &  \\
    South America & 3.62 & 7.38 & 2.76 &  & 3.46 &  & 5.25 &  \\
    Pacific Islands & 2.76 & 5.66 & 1.98 &  & 1.77 &  & 1.74 &  \\ \hline
    \end{tabular}
\end{table}


\noindent \textbf{Discussion and limitations}
There appears to be a bias towards the white race and European and North American countries in reference to STEM careers. This overrepresentation could be due to the racial and country of citizenship backgrounds of Wikidata contributors. Underrepresentation of other race and continent groups could stem from different ways to define careers such as a scientist; an individual from one of these groups could practice science in a different way from what is expected in Eurocentric cultures and therefore go unrecorded. Finally, we note that we compare the demographics of all people in Wikidata against the current world demographics. Follow-up work should bridge this gap, e.g., by only considering people in Wikidata that are alive today.

\section{Conclusions and Future Work}
\label{sec:conclusion}

In this paper, we analyzed the race and country of citizenship bias in Wikidata in general and for scientists, software developers, and engineers. We compared Wikidata queries with external data to see the representation differences. We ran a table linking software to match individuals from a Black Scientists dataset to Wikidata. An overrepresentation and bias of the white race and Europrean and North American country of citizenships was found, along with the underrepresentation of other races and countries in most subcategories. 

Future work can continue running the table linking software on the other scientist datasets and increase representation with a minority bot.

\section*{Acknowledgment}
We thank the USC Information Sciences Institute NSF REU site.

 \bibliographystyle{splncs04}
 \bibliography{refs}

\end{document}